\documentclass[10pt,twocolumn,letterpaper]{article}

\usepackage[pagenumbers]{iccv}

\definecolor{iccvblue}{rgb}{0.21,0.49,0.74}
\usepackage[breaklinks,colorlinks,allcolors=iccvblue]{hyperref}

\title{CanonNet: Canonical Ordering and Curvature Learning for Point Cloud Analysis}

\author{Benjy Friedmann\\
Hebrew University of Jerusalem\\
Jerusalem, Israel\\
{\tt\small benjamin.friedmann@mail.huji.ac.il}
\and
Michael Werman\\
Hebrew University of Jerusalem\\
Jerusalem, Israel\\
{\tt\small michael.werman@mail.huji.ac.il}
}

\begin{document}

\maketitle


\begin{abstract}
Point cloud processing poses two fundamental challenges: establishing consistent point ordering and effectively learning fine-grained geometric features. Current architectures rely on complex operations that limit expressivity while struggling to capture detailed surface geometry. We present CanonNet, a lightweight neural network composed of two complementary components: (1) a preprocessing pipeline that creates a canonical point ordering and orientation, and (2) a geometric learning framework where networks learn from synthetic surfaces with precise curvature values. This modular approach eliminates the need for complex transformation-invariant architectures while effectively capturing local geometric properties. Our experiments demonstrate state-of-the-art performance in curvature estimation and competitive results in geometric descriptor tasks with significantly fewer parameters (\textbf{100X}) than comparable methods. CanonNet's efficiency makes it particularly suitable for real-world applications where computational resources are limited, demonstrating that mathematical preprocessing can effectively complement neural architectures for point cloud analysis. The code for the project is
publicly available \hyperlink{https://benjyfri.github.io/CanonNet/}{https://benjyfri.github.io/CanonNet/}. 
\end{abstract}
\vspace{-0.4cm}
    
\section{Introduction}
\label{sec:intro}

\begin{figure}[t]
  \centering
   \includegraphics[width=0.8\columnwidth]{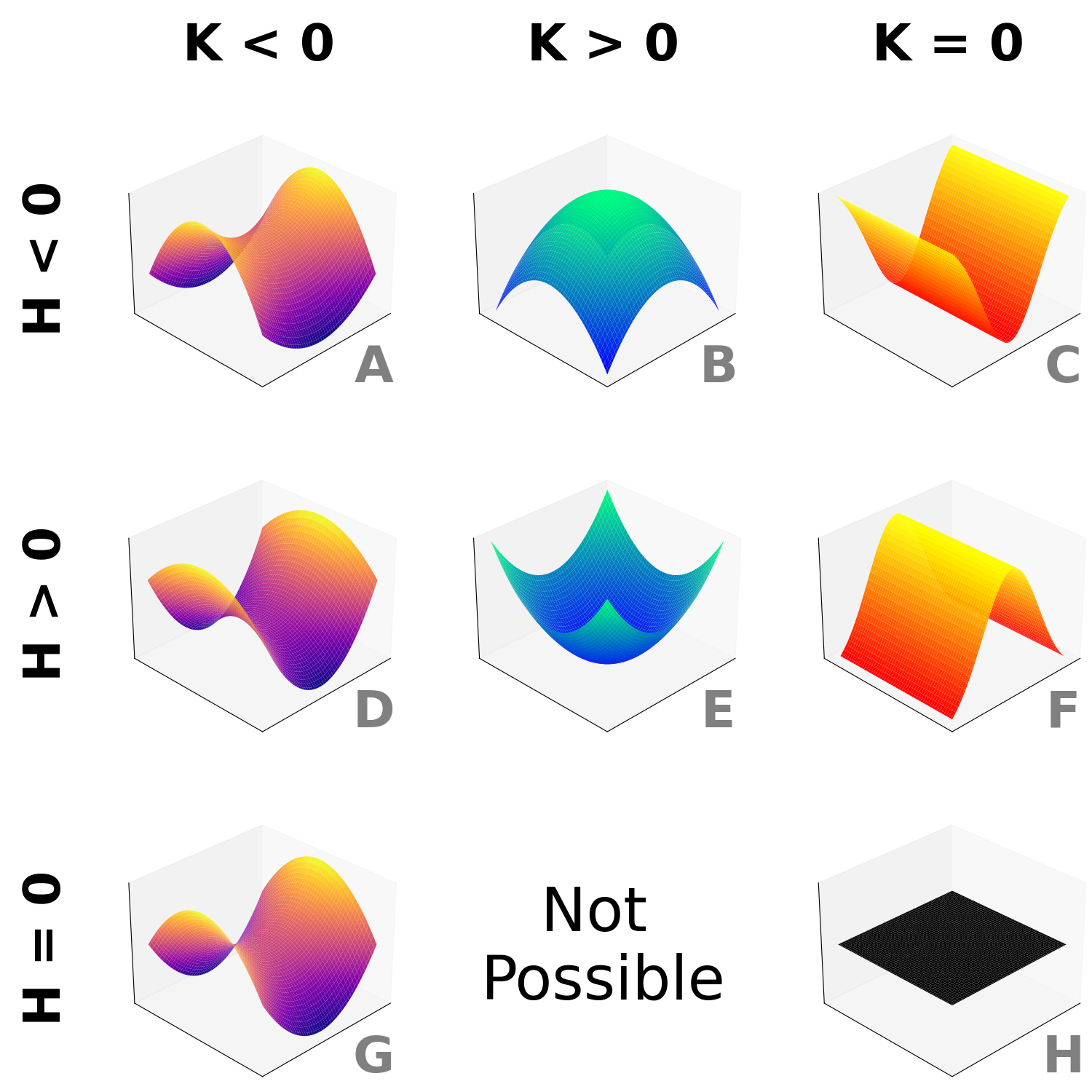}
    \vspace{-0.2cm}
   \caption{The different possible surfaces given the Gaussian (\textbf{K}) and mean (\textbf{H}) curvature as described in \cref{sec:training}. Note that up to rigid motion there are 4 different types of surfaces.}
   \vspace{-0.6cm}
   \label{fig:surfaces}
\end{figure}

Point clouds, which are unstructured collections of 3D points, have become fundamental to numerous applications including autonomous driving \cite{li2020deep}, robotics \cite{pomerleau2015review}, and medical imaging \cite{sitek2006tomographic}. While point clouds  capture detailed geometric information, their unstructured nature presents significant challenges for processing and analysis. Existing approaches have not fully resolved two critical challenges (1) establishing a consistent ordering of points and (2) effectively learning fine-grained geometric features.
In this paper, we present \emph{CanonNet}, a novel approach that establishes canonical point ordering and orientation while enhancing the learning of geometric features through synthetic data with \emph{precise} curvature annotations.

The unstructured nature of point clouds necessitates neural architectures that are permutation-invariant \cite{qi2017pointnet,qi2017pointnet++,guo2021pct, pan20213d, wang2019dynamic}. This is typically achieved through operations like  pooling, which aggregate point features regardless of their order. PointNet \cite{qi2017pointnet} pioneered the application of such operations in point cloud processing (\begin{eg}classification, segmentation    
\end{eg}), though similar permutation-invariant mechanisms had already been explored in Graph Neural Networks (GNNs) \cite{behler2007generalized,duvenaud2015convolutional}. While effective in ensuring order invariance, these symmetric aggregation functions inherently limit a network’s expressivity ~\cite{kondor2018covariant, de2020natural}, restricting its ability to capture fine-grained geometric relationships~\cite{joshi2023expressive}.

To enhance neural network expressivity when processing graphs, researchers have developed various positional encoding (PE) methods~\cite{grotschla2024benchmarking}. such as Laplacian-based, random walk-based, and other approaches. These techniques, particularly those using Laplacian eigenvectors, effectively encode structural properties~\cite{belkin2003laplacian, kreuzer2021rethinking, dwivedi2023benchmarking, maskey2022generalized}. However, in point cloud processing, PE has primarily been limited to Transformer-based architectures ~\cite{lai2022stratified, zhao2021point, qin2022geometric, pan20213d}, with some Transformer variants deliberately omitting it~\cite{guo2021pct}. We show that Laplacian PE can be harnessed in point cloud processing to achieve canonical order and orientation, eliminating the need for complex transformation-invariant architectures

The geometric properties inherent in point clouds constitute another valuable source of information that can be harnessed by neural architectures. 
Rather than relying solely on learned representations, various approaches exploit explicitly computed features such as normals
\cite{deng2018ppf, deng2018ppfnet, yuan2023egst}, angles \cite{deng2018ppf, deng2018ppfnet, yuan2023egst, qin2022geometric}, 
and pairwise distances \cite{deng2018ppf, deng2018ppfnet, yuan2023egst, qin2022geometric} as supplementary inputs to enhance model capabilities.
Among these geometric property approaches,  approximating curvature directly from point clouds via triangulation and supplying them as features to neural networks  has achieved significant performance gains~\cite{ran2022surface}.
In this context, a primary constraint is the limited availability of high-quality training data with accurate geometric annotations, which has restricted the evolution of learning-based approaches that can effectively utilize these geometric properties. Real-world point cloud datasets often lack precise geometric ground truth, making it difficult to train models that can reliably learn and interpret local surface properties. This limitation has particularly affected the development of approaches that aim to understand fine-grained geometric features.

\subsection*{Key Contributions}
\begin{enumerate}
\item 
A novel preprocessing pipeline that establishes both canonical point cloud ordering and canonical orientation, ensuring invariance to point permutations and rigid transformations respectively.

\item A synthetic data generation framework leveraging analytic surfaces with known curvatures, enabling unlimited training samples with precise geometric properties.
\item A lightweight neural network architecture that effectively learns local geometric features through curvature-based classification.
\end{enumerate}

\section{Related Works}
\label{sec:rel_work}

Point cloud processing presents unique challenges due to the irregular and unordered nature of the data. Our work advances this field through two key innovations: geometry-aware canonical ordering and curvature-based synthetic data generation. Here, we review relevant literature across three main areas that inform our approach.

\subsection{Deep Learning for Point Cloud Processing}
Point clouds' irregular and unordered nature presents fundamental challenges for deep learning approaches. While early methods converted point clouds to regular representations like voxel grids or 2D projections, these transformations introduced artifacts and lost geometric detail. PointNet~\cite{qi2017pointnet} introduced direct point cloud processing through point-wise MLPs and permutation-invariant pooling, though its point-independent processing limited local geometry capture. Building upon this foundation, PointNet++ ~\cite{qi2017pointnet++} introduced hierarchical sampling and grouping, enabling multi-scale feature learning at the cost of increased computational complexity.

Subsequent architectures enhanced local feature aggregation through various approaches. DGCNN ~\cite{wang2019dynamic} implemented neighborhood-aware processing through dynamic graph construction and edge convolutions. KPConv ~\cite{thomas2019kpconv} introduced learnable kernel points for geometry-adaptive convolutions. Recent transformer-based architectures, including Point Transformer ~\cite{zhao2021point} and PCT ~\cite{guo2021pct}, leverage self-attention mechanisms to model geometric relationships in local neighborhoods.

Despite these advances, current methods remain computationally intensive and struggle to fully capture underlying geometric structures, indicating the need for more efficient, geometry-aware approaches.

\subsection{Surface Geometry in Point Clouds} 
Surface geometry is important both as a self-contained task, such as surface normal reconstruction, and as part of the input to other models, where it provides key geometric information for tasks like registration and classification. Several learning-based methods use neural networks to predict local geometric properties directly from raw point clouds. PCPNet \cite{guerrero2018pcpnet} introduces a patch-based learning approach that encodes local point neighborhoods at multiple scales, enabling accurate surface normal and curvature estimation. Similarly, DeepFit \cite{ben2020deepfit} uses a surface fitting approach, where a neural network learns point-wise weights for weighted least-squares polynomial surface fitting. This method facilitates the extraction of normal vectors and other geometric properties, such as principal curvatures. Both methods leverage the PointNet \cite{qi2017pointnet} architecture, which provides a powerful framework for processing point clouds.

Incorporating geometric features such as distances, angles, and normals have been shown to improved performance in downstream tasks. Several works have demonstrated significant improvements by feeding these geometric features directly into their network architectures~\cite{rusu2008persistent,rusu2009fast,deng2018ppfnet,yuan2023egst,qin2022geometric}. Additionally, curvature estimation, commonly derived through triangulation, has been shown to boost performance when integrated into neural models \cite{ran2022surface}.

\begin{figure*}[t]
  \centering
   \includegraphics[width=0.9\textwidth]{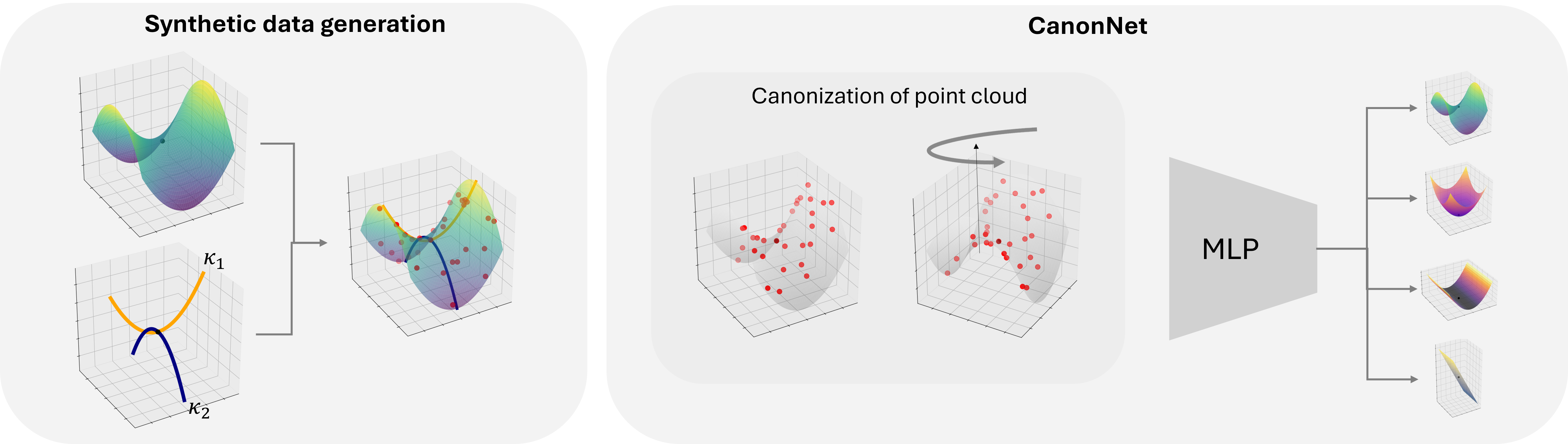}
   \caption{The complete CanonNet pipeline: Synthetic data generation (LHS): We sample points from analytically defined surfaces with known principal curvatures ${{\kappa}_1},{{\kappa}_2}$. Processing and classification (RHS): The point cloud is transformed into canonical orientation and processed by an MLP that performs supervised classification into four geometric surface types (Saddle, Parabolic, Valley, Plane) using ground truth curvature labels.}
   \vspace{-0.5cm}
   \label{fig:full_diag}
\end{figure*}
 
Our work builds on these approaches to efficiently estimate geometric properties (specifically curvature based) with minimal computational overhead. While previous methods rely on complex architectures to achieve invariance to permutation and 3D rigid transformations, our novel preprocessing pipeline establishes these invariances before the neural network.  This preprocessing is complemented by our analytical formulation that yields exact curvature measurements rather than conventional approximations. Together, these advances enable us to use small, expressive non-invariant architectures (\begin{eg}MLPs\end{eg}) that are significantly more parameter-efficient while maintaining competitive performance.
\subsection{Ordering and Orientation in Point Clouds}
Methods that transform a point clouds into a canonical orientation have emerged as an important direction in point cloud processing. Spatial Transformer Networks (STN) \cite{jaderberg2015spatial} introduced this concept in the 2D domain by predicting affine transformations to align images to a canonical form, enabling invariance to spatial transformations. Building on this foundation, PointNet \cite{qi2017pointnet} adapted the approach to 3D data with the T-Net module, which predicts rigid transformations to align point clouds before feature extraction, achieving invariance to geometric transformations. PCPNet \cite{guerrero2018pcpnet} enhanced the stability of this technique by constraining the spatial transformer to rotation-only transformations through quaternion representation, which both stabilized convergence and simplified the computation of inverse transformations for geometric properties. While these learned canonical representations improve performance, they do not provide theoretical guarantees of invariance to point permutation and rigid transformations.
The literature on canonical ordering of point clouds, however, remains limited. Most existing ordering approaches \cite{zheng2019pointcloud, yang2023self, dovrat2019learning} primarily address the challenge of identifying point importance for downsampling applications. These methods have demonstrated advantages over traditional techniques such as farthest point sampling and random sampling in downstream tasks including classification and registration, but they do not establish a truly canonical ordering scheme.

Our work builds upon these ideas of canonical orientation and geometric feature learning, while introducing novel contributions in point cloud processing through both canonical ordering (addressing point sequence permutation) and canonical orientation (ensuring consistent spatial alignment), alongside advancements in surface geometry learning. By combining our lightweight dual approach to canonical ordering and canonical orientation with curvature-based synthetic data, we enable more robust and geometrically meaningful point cloud processing. Notably, our method achieves this while requiring only small fraction of the computational resources needed by existing approaches, making it suitable for resource-constrained applications.

\section{Method}
We present a framework that combines differential geometry with deep learning to enable 
efficient point cloud processing. Our approach consists of two main components: 
(1) a preprocessing pipeline that establishes canonical point ordering and orientation, and (2) a training methodology utilizing synthetically 
generated surfaces with known geometric properties.

\subsection{Preprocessing Pipeline}
\label{sec:preprocessing}
This section details our preprocessing approach for creating consistent point ordering and orientation.
\begin{figure}[t]
  \centering
   \includegraphics[width=0.7\columnwidth]{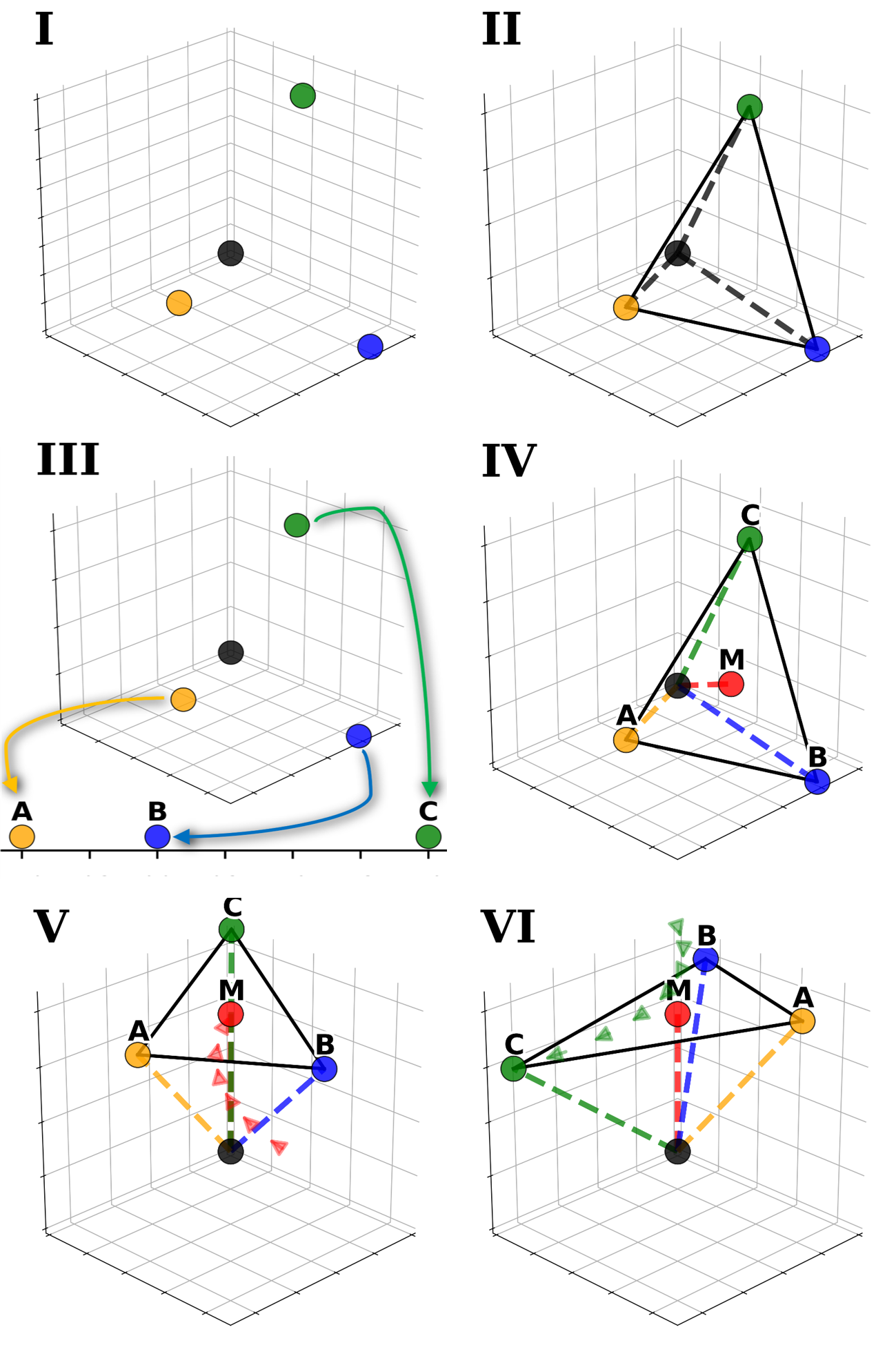}
    \vspace{-0.4cm}
   \caption{Illustration of the preprocessing pipeline for establishing canonical point cloud representation as described in \cref{sec:preprocessing}: \textbf{(I)} Input point cloud with arbitrary ordering and orientation. \textbf{(II)} Construction of fully connected graph with heat kernel weights and computation of normalized graph Laplacian. \textbf{(III)} Reordering points along a 1D axis based on Laplacian eigenvector values, ensuring consistency regardless of initial point indexing or spatial orientation and position. \textbf{(IV)} Identification of geometric landmarks: center of mass, 'M', and the point corresponding to the largest eigenvector value, 'A'. \textbf{(V)} First standardization rotation aligning center of mass with positive z-axis. \textbf{(VI)} Second standardization rotation placing 'A' in the XZ-plane with positive x-coordinate, completing the transformation pipeline that ensures both permutation and rigid-transformation invariance.}
   \vspace{-0.6cm}
   \label{fig:preprocess}
\end{figure}
\subsubsection{Graph Construction and Spectral Embedding}

Given a local patch from a 3D point cloud $\mX = \{ x_i \}_{i=1}^{N}, x_i \in \mathbb{R}^3$ or 
in matrix form $\quad \mX \in \mathbb{R}^{N \times 3}$, we aim to establish a consistent 
point ordering that is invariant to permutations and rigid transformations. To achieve 
this, we construct a fully connected, undirected graph $\gG = (\gV, \mW)$ to capture 
the local geometric relationships between points. The edge weights $\mW$ are defined 
by the heat kernel:
\begin{equation}
\mW_{ij} = \exp\left(-\frac{\|\vx_i - \vx_j\|_2^2}{t}\right) \quad \forall 
\vx_i, \vx_j \in \gV
\end{equation}
Here $t > 0$ is a temperature parameter controlling the locality of point interactions.
We compute the normalized graph Laplacian \cite{chung1997spectral}:

\begin{equation}
\mL = \Delta^{-1/2} \mW \Delta^{-1/2}
\end{equation}
where $\Delta \in \mathbb{R}^{N \times N}$ is the diagonal degree matrix with entries 
$\Delta_{ii} = \sum_{j=1}^N \mathbf{W}_{ij}$ for $i = 1, 2, \ldots, N$, and 
$\Delta_{ij} = 0$ for all $i \neq j$.

Next, we compute the eigenvector $\boldsymbol{\phi} \in \mathbb{R}^N$ corresponding to the smallest nonzero eigenvalue of $\mathbf{L}$~\cite{fiedler1973algebraic, fiedler1975property}. This eigenvector establishes a spectral embedding where each point $\mathbf{x}_i$ in the point cloud corresponds to the $i$-th component of $\boldsymbol{\phi}$, effectively projecting the 3D points onto a single line. As demonstrated by \cite{belkin2003laplacian}, this embedding minimizes the weighted sum $\sum_{i,j} \mathbf{W}_{ij}(\phi_i - \phi_j)^2$, ensuring that points with strong connections in the original 3D space remain close in the 1D embedding, thereby optimally preserving the local geometric structure.

\subsubsection{Canonical Ordering and Orientation}

As illustrated in \cref{fig:preprocess}, we establish a standardized point cloud representation through three complementary transformations that address ordering, position, and orientation. Together, these transformations ensure that geometrically equivalent shapes converge to identical representations regardless of their initial configurations.

The spectral embedding computed in the previous step provides the foundation for our canonical ordering. By arranging points according to their corresponding values in eigenvector $\boldsymbol{\phi}$, we define permutation $\sigma \in S_N$ where $\phi_{\sigma(1)} \leq \phi_{\sigma(2)} \leq \ldots \leq \phi_{\sigma(N)}$. The corresponding permutation matrix $\boldsymbol{\Pi}$, with elements $\Pi_{ij} = 1$ if $j = \sigma(i)$ and 0 otherwise, reorders the point cloud as:
\begin{equation}
\bar{\mathcal{X}} = \boldsymbol{\Pi}\mathcal{X}
\end{equation}
Positional normalization follows by aligning the center of mass with the z-axis. From the reordered point cloud, we compute the center of mass:
\begin{equation}
m={\frac{1}{N}}\sum_{i=1}^N \bar{\vx}_i
\end{equation}
We then determine a rotation matrix $\mathbf{R}_1$ that places $m$ at $(0, 0, \|m\|_2^2)$. Applying this rotation yields:
\begin{equation}
{\mathcal{Y}} = \mathbf{R}_1 \bar{\mathcal{X}}
\end{equation}
To ensure consistent orientation, we perform a final rotation about the z-axis based on a landmark point. Specifically, we identify a point $p_1=(x_1,y_1,z_1)$ in ${\mathcal{Y}}$ corresponding to the largest value in $\boldsymbol{\phi}$ and compute a rotation $\mathbf{R}_2$ that places $p_1$ in the $XZ$ plane with positive x-coordinate. This produces the final standardized representation:
\begin{equation}
{\mathcal{P}} = \mathbf{R}_2 {\mathcal{Y}}
\end{equation}
The complete transformation pipeline is:
\begin{equation}
{\mathcal{P}} = \mathbf{R}_2 \mathbf{R}_1 \mathbf{{\Pi}} {\mathcal{X}}
\end{equation}

\subsubsection{Invariance Properties}

A key advantage of our preprocessing pipeline is its theoretical guarantees of invariance to both point permutations and rigid transformations. We formally establish these properties below.
\begin{theorem}
The proposed preprocessing pipeline is invariant to point permutation and rigid transformation.
\begin{proof}
Let $\mX \in \mathbb{R}^{N \times 3}$ be a 3D point cloud, $\mP \in \mathbb{R}^{N \times N}$ a permutation matrix, $\mR \in \mathbb{R}^{3\times3}$ a rotation matrix, and $\vt \in \mathbb{R}^{3}$ a translation vector. We define the transformed point cloud as $\tilde{\mX} = \mR\mX + \vt$.

\smallskip\noindent
\textbf{Rigid Transformation Invariance:} 
For any pair of points $\vx_i, \vx_j \in \mX$, their pairwise distance remains unchanged under rigid transformation:
\begin{equation}
\begin{split}
\|\tilde{\vx}_i - \tilde{\vx}_j\| &= \|(\mR\vx_i + \vt) - (\mR\vx_j + \vt)\| \\
&= \|\mR(\vx_i - \vx_j)\| = \|\vx_i - \vx_j\|
\end{split}
\end{equation}

Thus, the weight matrix $\mW_{i,j} = \exp\left(-\frac{\|\vx_i - \vx_j\|_2^2}{t}\right)$ remains invariant, leading to an identical Laplacian matrix.

\smallskip\noindent
\textbf{Point Permutation Invariance:} Let $\boldsymbol{\phi}$ be the eigenvector corresponding to the smallest non-zero eigenvalue $\lambda$ of the Laplacian matrix $\mL$. We assume that the multiplicity of  $\lambda$ is $1$ (true for typical point distributions, as infinitesimal perturbations break degeneracy), ensuring that $\boldsymbol{\phi}$ is unique up to scaling.
For a permuted point cloud $\bar{\mX} = \mP\mX$, with permutation matrix $\mP$, the weight matrix transforms as $\bar{\mW} = \mP\mW\mP^{-1}$, resulting in a similarity-transformed Laplacian $\bar{\mL} = \mP\mL\mP^{-1}$. 

For the eigenvector $\boldsymbol{\phi}$ of $\mL$ with eigenvalue $\lambda$, we have:
\begin{align}
\bar{\mL}(\mP\boldsymbol{\phi}) = \mP(\mL\boldsymbol{\phi}) = \mP(\lambda\boldsymbol{\phi}) = \lambda(\mP\boldsymbol{\phi})
\end{align}
Therefore, $(\mP\boldsymbol{\phi})$ is an eigenvector of $\bar{\mL}$ with eigenvalue $\lambda$. Since $\lambda$ has multiplicity $1$, eigenvectors for the original and permuted Laplacians differ only by the permutation $\mP$ and scaling.

To establish a canonical ordering, we first normalize $\boldsymbol{\phi}$ so its largest-magnitude entry is positive, resolving sign ambiguity. We then permute the points according to these normalized eigenvector values, yielding a point ordering invariant to initial permutations.
\end{proof}
\end{theorem}

These invariance properties enable our lightweight MLP architecture to focus exclusively on learning geometric features, without needing complex structures to handle permutation and transformation equivariance.

These invariance properties ensure that our preprocessing pipeline produces consistent results regardless of the initial point ordering or orientation of the input point cloud. This theoretical guarantee enables our lightweight MLP architecture to focus on learning the geometric properties of the surface rather than accounting for permutation and transformation variations.

\subsection{Training}
\label{sec:training}
This section outlines our geometric learning framework. We first present our synthetic 
data generation approach with controlled curvature properties. Then we describe our geometric feature extraction process and the design of our parameter-efficient network for surface classification.
\subsubsection{Synthetic Data Generation}
\label{sec:synthetic_data_generation}
Surface curvature quantifies how a surface deviates from being flat at a point. The principal curvatures $\kappa_1$ and $\kappa_2$  represent the maximum and minimum bending of the surface. From these, we derive the Gaussian curvature $K=\kappa_1\kappa_2$ and the mean curvature $H=\kappa_1+\kappa_2$.
For our dataset, we generate quadratic surfaces with analytically tractable curvature properties:
\begin{equation}
    z = f(x,y) = ax^2 + by^2 + cxy + dx + ey
\end{equation}
 Among possible sampling methods, we chose to sample coefficients $(a,b,c,d,e)$, which proved sufficient to create surfaces with the full range of desired curvature characteristics. These surfaces are classified into one of four categories: $\mathcal{C} = \{\text{plane}, \text{parabolic}, \text{valley}, \text{saddle}\}$ based on its curvature signature at the origin. 

To ensure an unbiased dataset, we sample each class separately with equal representation.
For each generated surface, we uniformly sample points within the region 
 $[-0.5 , 0.5] \times [-0.5 , 0.5]$ to create our synthetic dataset.

\subsubsection{Feature Extraction and Network Design}
\label{sec:feature_extraction_and_network_design}
Our training process employs a supervised learning approach using synthetic point cloud data, generated as described in the previous section. This dataset serves as the foundation for training a lightweight Multi-Layer Perceptron (MLP) designed to classify different surface types based on geometric features.
The input to the MLP consists of the preprocessed point cloud ${\mathcal{P}}$, along with a set of second-order polynomial features derived from the coordinates of each point (\begin{eg} \(x^2\), \(y^2\), \(xy\)\end{eg}).
These polynomial terms encode local curvature variations, enabling the model to capture higher-order geometric relationships within the point cloud. By leveraging these enhanced representations, the network improves its ability to differentiate between surface classes with varying curvature characteristics.
The network is trained to classify each input point cloud ${\mathcal{P}}$ into one of four fundamental surface categories:
\begin{enumerate}
    \item \textbf{Plane:} A flat surface characterized by zero Gaussian and mean curvatures.
    \item \textbf{Parabolic:} A convex surface with positive Gaussian curvature (\begin{eg}spheres, domes\end{eg}). 
    \item \textbf{Valley:} A surface with zero Gaussian curvature and nonzero mean curvature (\begin{eg}a cylinder\end{eg}).
    \item \textbf{Saddle:} A hyperbolic surface with negative Gaussian curvature.
\end{enumerate}
These four surface types represent fundamental geometric structures commonly encountered in real-world point cloud data. Their classification is critical for downstream tasks such as shape reconstruction and geometric reasoning. \cref{fig:surfaces} provides visual illustrations of these surfaces, highlighting their distinct curvature properties.

\section{Experiments}
\label{sec:experiments}

\begin{table}[t]
  \scriptsize
  
  \centering
  \resizebox{\linewidth}{!}{
  \begin{tabular}{@{}lcccl@{}}
    \toprule
    \multicolumn{5}{c}{\textbf{Results on PCPNET Dataset}} \\
    \midrule
    Method & $D_K \downarrow$ & $D_H \downarrow$ & \#Params (M) & \#Points \\
    \midrule
    PCPNET \cite{guerrero2018pcpnet} & 6.88 & 1.91 & 22 & 500 \\
    DeepFit \cite{ben2020deepfit} & 0.56 & 0.67 & 3.5 & 128 \\
    \textbf{CanonNet}  & 8.2 & \textbf{0.4} & \textbf{0.03} & \textbf{20} \\
    \midrule
    \multicolumn{5}{c}{\textbf{Results on Synthetic Dataset}} \\
    \midrule
    \textbf{CanonNet}  & 0.97 & \textbf{0.14} & \textbf{0.03} & \textbf{20} \\
    \bottomrule
  \end{tabular}}
  \vspace{-0.3cm}
  \caption{Comparison of Gaussian ($D_K$) and mean ($D_H$) curvature estimation errors (RMSE) across different methods and datasets. CanonNet uses significantly fewer parameters and points compared to other methods while achieving competitive performance.}
  \label{tab:PCPNET_dataset}
\end{table}
\begin{table}[t]
  \centering
  \resizebox{\linewidth}{!}{
  \begin{tabular}{@{}lccc@{}}
    \toprule
    Method               & Param. (Mb)  & In-Domain (\%) & Cross-Domain (\%) \\
    \midrule
    FPFH \cite{choy2019fully}           & -            & 35.9          & 22.1            \\
    SHOT \cite{tombari2010unique}          & -            & 23.8          & 61.1            \\
    3DMatch \cite{zeng20173dmatch}        & 13.40        & 59.6          & 16.9            \\
    CGF \cite{khoury2017learning}          & 1.86         & 58.2          & 20.2            \\
    PerfectMatch\cite{gojcic2019perfect}    & 3.26         & 94.7          & 79.0            \\
    FCGF \cite{choy2019fully}             & 33.48        & 95.2          & 16.1            \\
    D3Feat (rand) \cite{bai2020d3feat}    & 13.42        & 95.3          & 26.2            \\
    LMVD \cite{li2020end}            & 2.66         & 97.5          & 79.9            \\
    SpinNet \cite{ao2021spinnet}         & 2.16         & 97.6          & 92.8            \\
    \textbf{CanonNet}                 & \textbf{0.03}          & -          & 65.7            \\
    \bottomrule
  \end{tabular}}
  \vspace{-0.3cm}
  \caption[Comparison of methods by parameter count and average False Match Rate]
  {Comparison of methods by parameter count and average False Match Rate (FMR). In-Domain shows performance on 3DMatch when trained on 3DMatch. Cross-Domain shows performance on unseen datasets. CanonNet has no In-Domain value as it was trained exclusively on our synthetic dataset, with its Cross-Domain value representing performance on 3DMatch. Notably, CanonNet achieves competitive Cross-Domain performance with only a fraction of the parameters used by other methods. Performance values reported by \cite{ao2021spinnet}.}
  \vspace{-0.4cm}
  \label{tab:fmr}
\end{table}

We evaluate CanonNet's performance on point cloud analysis tasks including Gaussian and mean curvature estimation and geometric descriptor-based retrieval. Our experiments demonstrate the framework's effectiveness in capturing local geometric features with minimal computational resources while using only synthetic training data.

\subsection{Gaussian and Mean Curvature Estimation}
\label{sec:gaussian_and_mean_curvature_estimation}
\subsubsection{Experimental Setup}

We evaluate CanonNet on the PCPNet dataset \cite{guerrero2018pcpnet}, which consists of point clouds sampled from various 3D shapes with ground truth normals and curvatures. Unlike previous approaches that train directly on this dataset, we train our model exclusively on synthetic quadratic surfaces as described in \cref{sec:training}. We employ a lightweight MLP architecture with only 0.03M parameters, processing small local neighborhoods of just 20 points after canonical ordering.

It is important to note that while methods like PCPNet \cite{guerrero2018pcpnet} and DeepFit \cite{ben2020deepfit} estimate principal curvatures directly, our work focuses on the local geometry which requires values that are agnostic to sign. Therefore, we chose to estimate gaussian curvature ($K = \kappa_1\kappa_2$) and absolute mean curvature ($|H| = |\frac{(\kappa_1+\kappa_2)}{2}|$), which are invariant to the choice of normal direction.

For our curvature estimation task, we augment our model's output to include both the surface type classification (resulting in 4 surface categories) and the gaussian and mean curvature values. This integrated approach enables our network to develop a deeper understanding of fundamental surface geometries, which in turn leads to more accurate curvature estimation. 

For comparison, we include state-of-the-art methods PCPNet \cite{guerrero2018pcpnet} and DeepFit \cite{ben2020deepfit}, both of which were specifically designed for normal and curvature estimation and trained directly on the PCPNet dataset. These methods use significantly larger patch sizes (500 and 128 points, respectively) and more complex architectures (22M and 3.5M parameters, respectively).

\subsubsection{Evaluation Metric}

We evaluate curvature estimation performance using the rectified error metric, which is calculated as:

\begin{equation}
D_K = \frac{|{K - K_{GT}}|}{\max\{|K_{GT}|, 1\}}
\end{equation}

\begin{equation}
D_H = \frac{|{H - H_{GT}}|}{\max\{|H_{GT}|, 1\}}
\end{equation}
where $K$ and $H$ are the predicted Gaussian and mean curvatures, and $K_{GT}$ and $H_{GT}$ are the ground truth values. The final error metrics are reported as the root mean square error (RMSE) of these normalized differences. This metric normalizes the error by the maximum of the absolute ground truth value and 1.0, ensuring stable evaluation across regions with different curvature magnitudes.

\subsubsection{Results and Analysis}

\cref{tab:PCPNET_dataset} presents the quantitative results for Gaussian and mean curvature estimation errors on the PCPNet dataset. Despite not being trained on this dataset and using dramatically smaller patch sizes, CanonNet achieves competitive performance. Specifically, while our model shows higher Gaussian curvature error (8.2) when directly applied to the PCPNet dataset, it achieves state-of-the-art performance on mean curvature estimation (0.4), outperforming both PCPNet (1.91) and DeepFit (0.67).

On synthetic data with analytically defined curvature values, CanonNet achieves exceptional performance with Gaussian curvature error of only 0.97 and mean curvature error of just 0.14. 

The most striking aspect of these results is the parameter efficiency of our approach. CanonNet requires just 0.03M parameters, which is approximately \textbf{700}$\mathbf{\boldsymbol{\times}}$  smaller than PCPNet and \textbf{116}$\mathbf{\boldsymbol{\times}}$  smaller than DeepFit. This dramatic reduction in model size is achieved through our canonical preprocessing pipeline, which eliminates the need for complex architectures to achieve permutation and rotation invariance.

Additionally, CanonNet operates on much smaller local neighborhoods (20 points) compared to PCPNet (500 points) and DeepFit (128 points), significantly reducing computational overhead during inference. This makes our approach particularly suitable for resource-constrained applications where memory and processing power are limited.

\subsection{Geometric Descriptor Retrieval}
\label{sec:geometric_descriptor_retrieval}
\subsubsection{Experimental Setup}

It is important to note that CanonNet was not explicitly trained to be a geometric descriptor. Instead, we leverage the geometric understanding it develops through curvature estimation and surface classification to serve as an implicit descriptor. This is a significant distinction from other methods which are specifically designed and trained for descriptor-based matching tasks.

To create a more robust descriptor, we apply CanonNet to patches at multiple resolutions (created by progressively downsampling the point cloud). We then concatenate these multi-resolution outputs to form our final descriptor.

To evaluate CanonNet's effectiveness as a geometric descriptor for point cloud registration, we assess its performance using the Feature Match Recall (FMR)~\cite{deng2018ppf} metric on standard benchmark \cite{zeng20173dmatch}. FMR measures the percentage of point pairs with ground truth overlap that are correctly matched based on their feature descriptors.
For our evaluation, we focus specifically on mutual nearest neighbors (best-buddies) in the embedding space. The FMR is calculated as follows:
\begin{equation}
\text{FMR} = \frac{1}{N} \sum \mathbbm{1}[\|x_i - T\cdot y_j\| < \tau_1]
\end{equation}
where $N$ is the number of ground-truth corresponding point pairs, $x_i$ and $y_j$ are the corresponding best-buddy point pairs identified through mutual nearest-neighbor search in the descriptor space, $T$ is the ground-truth transformation, and $\tau_1$ is the Euclidean distance threshold to determine whether the matching pair is correct. The indicator function $\mathbbm{1}[{\cdot}]$ equals 1 when the condition is satisfied and 0 otherwise.

Following standard practice, we evaluate both on \emph{"In-Domain"} data (similar to the training distribution) and \emph{"Cross-Domain"} data (novel geometries not represented in the training set). However, unlike competing methods that train directly on real-world datasets, CanonNet is trained exclusively on our synthetic surface dataset.

We compare against a range of traditional handcrafted descriptors (FPFH \cite{rusu2009fast}, SHOT \cite{tombari2010unique}) and learning-based approaches (3DMatch \cite{zeng20173dmatch}, CGF \cite{khoury2017learning}, PerfectMatch \cite{gojcic2019perfect}, FCGF \cite{choy2019fully}, D3Feat \cite{bai2020d3feat}, LMVD \cite{li2020end}, and SpinNet \cite{ao2021spinnet}). These methods vary substantially in parameter count and architectural complexity.

\subsubsection{Results and Analysis}

\cref{tab:fmr} presents the comparison between CanonNet and existing methods in terms of parameter count and Feature Match Recall (FMR) percentages. Our approach achieves 65.7\% FMR on unseen data, which is competitive with several established methods despite being trained solely on synthetic data and using significantly smaller local neighborhoods.

While specialized methods like SpinNet achieve higher FMR (92.8\% on unseen data), they require 21.6$\times$ more parameters than CanonNet. Similarly, LMVD achieves 79.9\% FMR but requires 26.6$\times$ more parameters. This highlights the favorable trade-off that CanonNet offers between performance and computational efficiency.
When applying these descriptors to point cloud registration tasks, our approach still maintains practical efficiency. Although CanonNet may have a lower inlier rate compared to more complex models, RANSAC-based registration using our descriptors requires only 5 samplings on average to find correct inlier correspondences. This minimal RANSAC overhead is easily offset by our method's significant speed advantage in descriptor generation.
The most significant advantage of our approach is its minimal parameter count of just 0.03Mb, making it by far the most compact model among all compared methods. This extreme parameter efficiency is achieved through our canonical preprocessing pipeline, which eliminates the need for complex architectures to handle point permutations and rigid transformations.

Building on this small parameter footprint, CanonNet further enhances computational efficiency by processing  300 total points per patch (across all resolution levels), while competing methods typically require 1000-2000 points. This dramatic reduction in both model size and input size collectively contributes to minimizing computational requirements, making our approach particularly suitable for applications with real-time processing needs or limited resources. For context, our implementation generates descriptors approximately 30 times faster than SpinNet on a standard GPU, demonstrating the significant performance advantages of our lightweight architecture.

\section{Ablation Studies}
\label{sec:ablation}

\begin{figure*}[h!]
\centering

\begin{tabular}{c@{}c@{}c@{}}
\includegraphics[width=0.33\textwidth]{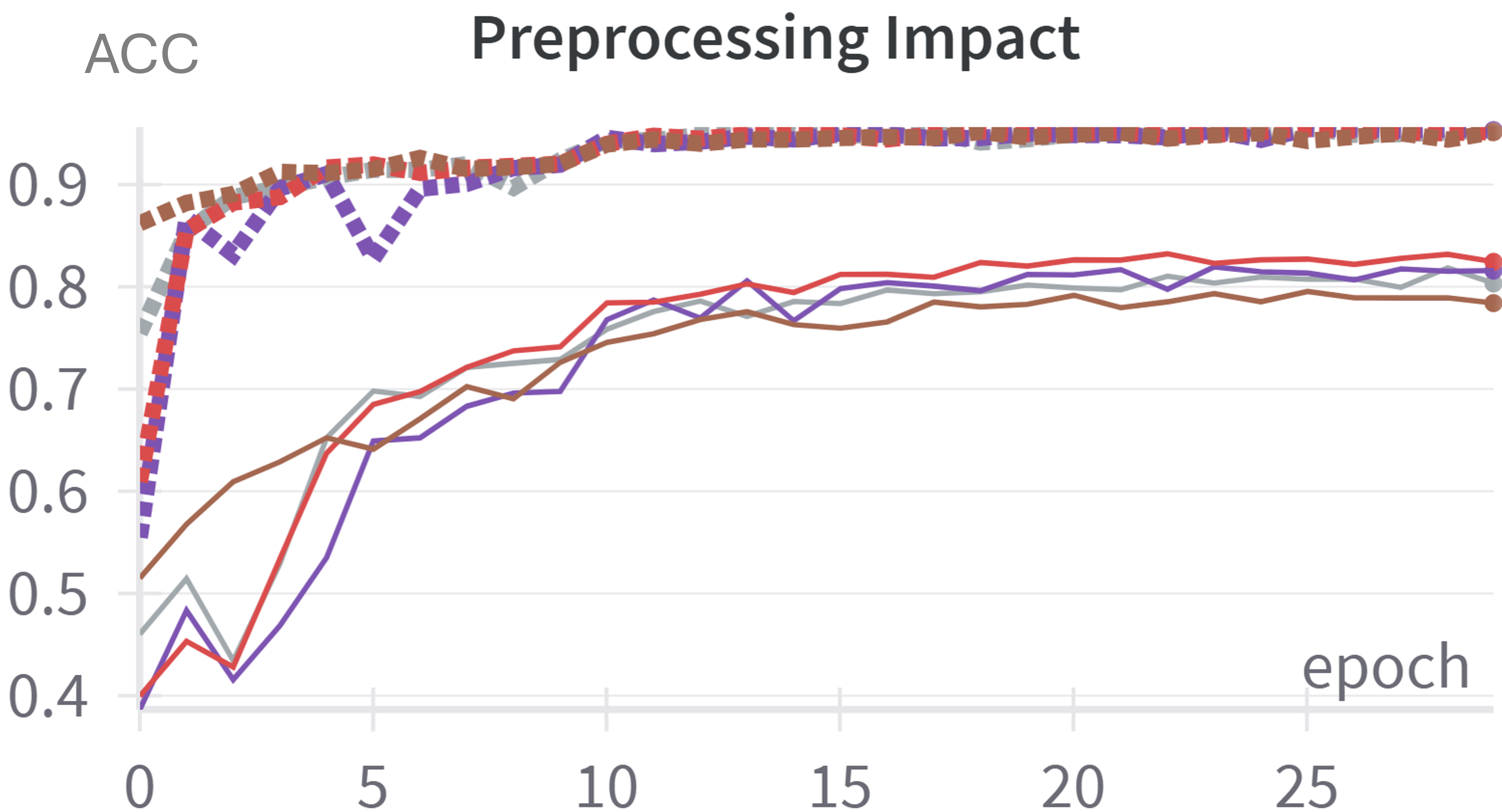}  & 
\includegraphics[width=0.33\textwidth]{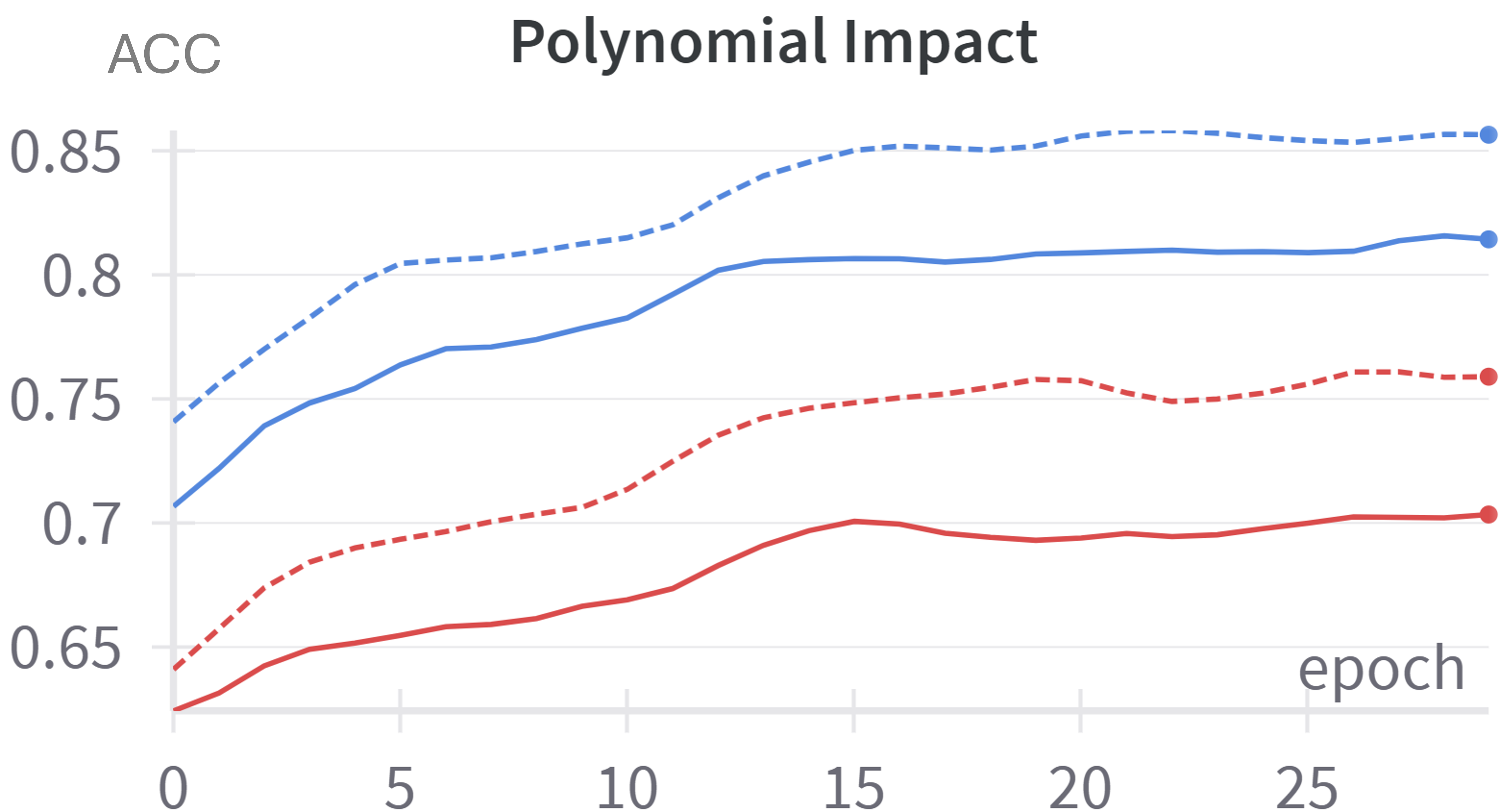} & 
\includegraphics[width=0.33\textwidth]{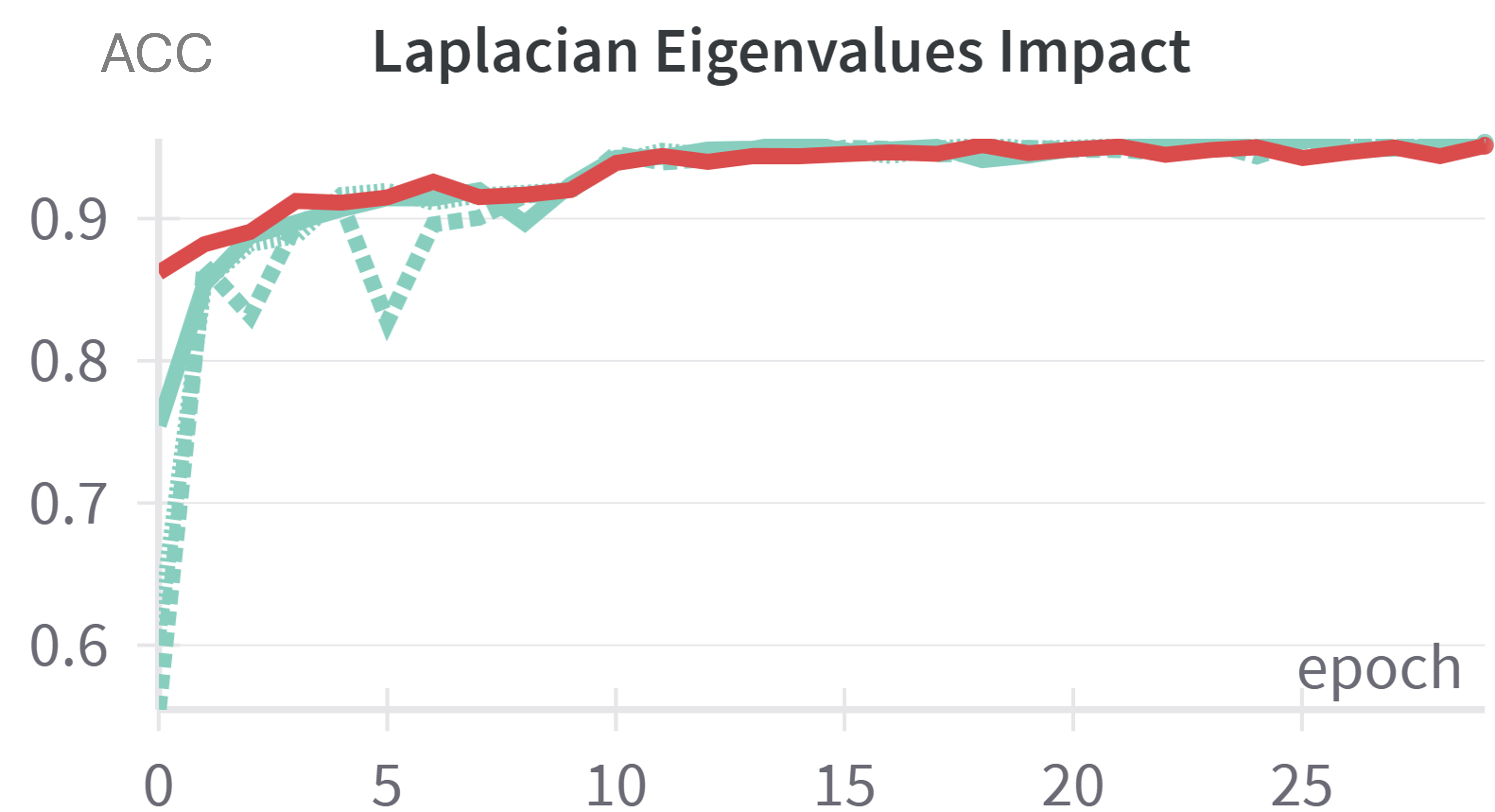} \\
\textbf{(a)} & \textbf{(b)} & \textbf{(c)}
\end{tabular}
\vskip -0.1in
\caption{\textbf{(a)} Impact of canonical preprocessing pipeline, showing consistent 10-17\% accuracy improvements across architectures and noise levels (solid: baseline, dashed: with preprocessing).
\textbf{(b)} Effect of second-degree polynomial features, yielding approximately 5\% accuracy improvement across all tested architectures (solid: baseline, dashed: with polynomial features).
\textbf{(c)} Impact of Laplacian eigenvalues, showing minimal differences (±0.2\%), suggesting geometric information is already well-captured by existing features.}

\label{fig:ablation_study}
\vspace{-0.4cm}
\end{figure*}

To systematically evaluate the contribution of individual components within the CanonNet architecture, we conducted a comprehensive series of ablation experiments. These investigations quantify the impact of four critical elements: (1) graph Laplacian formulations for canonical ordering, (2) the preprocessing pipeline's effect on performance, (3) second-degree polynomial features, and (4) Laplacian eigenvalues as supplementary input features.

\subsection{Graph Laplacian Selection}
\label{sec:graph_laplacian_selection}
\begin{table}[t]
  \centering
  \resizebox{\linewidth}{!}{
  \begin{tabular}{@{}lcccccc@{}}
    \toprule
    \textbf{Temp. / Noise} & 0 & 1\% & 3\% & 5\% & 7\% & 10\% \\
    \midrule
    t=0.5 (Norm) & 100 & 83 & 62.76 & 49.86 & 41.33 & 33.19 \\
    t=0.5 & 100 & 83.19 & 63 & 50.90 & 43.19 & 34.76 \\
    t=1 (Norm) & 100 & 83 & 63.38 & 51.05 & 42.57 & 34.57 \\
    t=1 & 100 & 82.05 & 62.19 & 50.67 & 42.38 & 34.24 \\
    t=2 (Norm) & 100 & 83.48 & 63.19 & 51 & 43.43 & 34.76 \\
    t=2 & 100 & 81.90 & 61.38 & 49.86 & 42.24 & 33.48 \\
    t=5 (Norm) & 100 & 83.19 & 62.81 & 51.33 & 43.43 & 34.71 \\
    t=5 & 100 & 81.95 & 61.76 & 49.67 & 41.29 & 33.48 \\
    \bottomrule
  \end{tabular}}
  \vspace{-0.2cm}
  \caption{Effect of Laplacian normalization and heat kernel temparature on robustness to noise. The values represent the percent of points in the same ordering after perturbations and applying our preprocessing pipeline.}
  \vspace{-0.4cm}
  \label{tab:noise_robustness}
\end{table}

Results in~\cref{tab:noise_robustness}, show normalized graph Laplacians generally perform slightly better than unnormalized versions when exposed to noise. This advantage remains consistent across all noise levels tested. Since different temperature settings produced nearly identical results, we selected $t=1$ for our normalized formulation implementation.


\subsection{Impact of Canonical Preprocessing Pipeline}
\label{sec:canonical_preprocessing_pipeline}
Our canonical preprocessing pipeline significantly improved classification performance across all tested architectures. 
\cref{fig:ablation_study} (a) shows consistent accuracy improvements of approximately 10\% when using canonical ordering and orientation, 
demonstrating that our approach effectively addresses permutation and rotation invariance challenges. 
This allows models to focus on learning geometric features rather than transformation variations.

When subjected to Gaussian noise perturbations, the performance gap widened to nearly 15\%, highlighting the pipeline's importance for establishing robust geometric feature learning

\subsection{Second-Degree Polynomial Features}
\label{sec:second_degree_polynomial_features}
We tested whether adding second-degree polynomial terms as input features would improve the model's ability to capture curvature information efficiently.

As shown in~\cref{fig:ablation_study} (b), these features consistently improved classification accuracy by approximately 5\% across all model configurations, with deeper networks showing more pronounced benefits.

These results confirm our hypothesis that explicit polynomial terms help models learn surface curvature characteristics by providing a mathematical basis aligned with differential geometry, allowing even simple architectures to distinguish surface types without extensive computational resources.


\subsection{Laplacian Eigenvalues as Input Features}
\label{sec:laplacian_eigenvalues_as_input_feaures}
We tested whether adding Laplacian eigenvalues as input features would improve geometric understanding, given their theoretical connection to intrinsic surface properties.

As shown in~\cref{fig:ablation_study} (c), this approach did not significantly impact performance, with classification accuracy changing by only ±0.2\% across all tested architectures.

This suggests our existing feature representation (3D coordinates and second-degree polynomial terms) already captures the essential geometric information in the Laplacian spectrum, making the additional computational cost unjustified.

\section{Conclusion}
We presented CanonNet, a lightweight neural network for point cloud analysis that achieves permutation and rotation invariance through a novel preprocessing pipeline. Combining canonical ordering and orientation with curvature-based synthetic data generation, our approach demonstrates competitive performance while requiring fewer parameters (0.03M) and smaller patch sizes (20 points) than state-of-the-art methods.

Results confirm CanonNet achieves state-of-the-art mean curvature estimation accuracy on the PCPNet dataset and competitive feature match recall, all with a parameter footprint orders of magnitude smaller than comparable approaches. This efficiency makes CanonNet suitable for resource-constrained applications, establishing a foundation for more efficient point cloud processing across numerous domains.

\clearpage
{
    \small
    \bibliographystyle{ieeenat_fullname}
    \bibliography{main}
}

\end{document}